\documentclass{article}
\usepackage{amsmath,graphicx,mlspconf}
\usepackage{xcolor}
\usepackage{multirow}

\usepackage{caption}
\usepackage{booktabs} % For better table lines
\usepackage{tcolorbox}

\usepackage{tikz}
\usepackage{pgfplots, pgfplotstable}
\usepackage{caption}
\usepackage{subcaption}
\usetikzlibrary{patterns}

\usepackage{amssymb}
\usepackage{amsmath}

\newcommand{\content}[1]{{#1}}
\copyrightnotice{U.S.\ Government work not protected by U.S.\ copyright}

\copyrightnotice{979-8-3503-2411-2/25/\$31.00 {\copyright}2025 Crown}

\copyrightnotice{979-8-3503-2411-2/25/\$31.00 {\copyright}2025 European Union}

\copyrightnotice{979-8-3503-2411-2/25/\$31.00 {\copyright}2025 IEEE}

\toappear{2025 IEEE International Workshop on Machine Learning for Signal Processing, Aug.\ 31-- Sep.\ 3, 2025, Istanbul, Turkey}

\title{Self-Supervised Learning at the Edge: The Cost of Labeling
}

\names{%
   Roberto Pereira$^{\star}$
   \qquad Fernanda Famá$^{\star}$
   \qquad Asal  Rangrazi$^{\dagger}$
   \qquad Marco Miozzo$^{\star}$
   }
  {Charalampos Kalalas$^{\star}$
   \qquad Paolo Dini$^{\star}$
   \thanks{This work has been partially funded by the grant CHIST-ERA-20-SICT-004 (SONATA) by PCI2021-122043-2A/AEI/10.13039/501100011033.
   \\
   $^{\dagger}$Work conducted while affiliated with Sustainable AI, CTTC.
    }% <-this % stops a space
   }
\address{%
   $^{\star}$Sustainable AI, CTTC/CERCA, Spain
   \\
   $^{\dagger}$Barcelona Supercomputing Center (BSC), Spain%
  \\
  $^{\star}$\{name.surname\}@cttc.es \qquad
$^{\dagger}$\{name.surname\}@bsc.es
}

\usepackage{url}

\begin{document}
%\ninept

% \setlength{\abovedisplayskip}{0.5pt}% template: 8,75 pt
% \setlength{\belowdisplayskip}{6.5pt}% template 8,75 pt

\maketitle

\begin{abstract}
Contrastive learning (CL) has recently emerged as an alternative to traditional supervised machine learning solutions by enabling rich representations from unstructured and unlabeled data. However, CL and, more broadly, self-supervised learning (SSL) methods often demand a large amount of data and computational resources, posing challenges for deployment on resource-constrained edge devices. In this work, we explore the feasibility and efficiency of SSL techniques for edge-based learning, focusing on trade-offs between model performance and energy efficiency. In particular, we analyze how different SSL techniques adapt to limited computational, data, and energy budgets, evaluating their effectiveness in learning robust representations under resource-constrained settings. 
Moreover, we also consider the energy costs involved in labeling data and assess how semi-supervised learning may assist in reducing the overall energy consumed to train CL models. 
Through extensive experiments, we demonstrate that tailored SSL strategies can achieve competitive performance while reducing resource consumption by up to $4\times$, underscoring their potential for energy-efficient learning at the edge.
\end{abstract}

\section{Introduction}
\label{sec:intro}

With the increasing number of edge devices,  in domains such as healthcare, autonomous systems, and environmental monitoring, there is a growing demand for intelligent algorithms that can operate efficiently and locally \cite{wang2020convergence}.  These devices generate large amounts of unlabeled data, making traditional supervised learning approaches at the edge impractical or costly due to the need for manual annotation.

Self-supervised learning (SSL) has emerged as a powerful alternative to traditional supervised approaches by leveraging intrinsic structure in the data to learn useful representations without explicit labels. Among SSL methods, contrastive learning (CL) techniques such as SimCLR \cite{simclr_chen2020simple} 
have demonstrated remarkable effectiveness in image and signal domains. Nonetheless, despite recent advances in making these methods more efficient--e.g., reducing the need for large batch sizes \cite{he2020momentum} and negative pairs \cite{simsiam_chen2021exploring, byol_grill2020bootstrap}--SSL models still often demand substantial computational resources and long training durations. These requirements pose significant challenges for deployment in edge scenarios, where memory, processing, data, and energy availability are inherently limited.

Existing literature often considers supervised and self-supervised learning as two disjoint paradigms, with models either assuming full access to labeled data or no access at all. Nonetheless, partial labeling may be achievable in certain contexts \cite{seimcl_yang2022class} at the cost of energy and manual labor; factors %Both of which are 
often neglected in performance evaluations.
This is especially relevant in energy-constrained edge environments (e.g., Internet of Things), where data availability may be scarce locally, subject to continuous change, and prone to abrupt drift \cite{dutta2021tinyml}. In such scenarios, the trade-off between labeling effort and model performance must be carefully balanced.  

Surprisingly, the energy cost of label acquisition--{either via manual annotation at the edge or by transmitting the raw data to a server %over a 
using a communication link}--and its impact on total training energy efficiency remain widely underexplored and are often not considered when studying the efficiency and scalability of machine learning solutions. In this work, we investigate the energy-performance trade-offs of supervised and self-supervised CL, accounting for both training energy and the energy required for label acquisition. Additionally, we explore semi-supervised learning scenarios where a small portion of the data is labeled and the remaining is not.
\content{We quantify the energy costs of training supervised, semi- and self-supervised CL solutions in different data regimes, while also considering the cost of data labeling.  Our findings reveal that label acquisition may consume
% a comparable or greater
twice the amount of energy consumed for training, thereby challenging assumptions about the efficiency of supervised learning at the edge.}
The main contributions of this paper can be summarized as:
\vspace{-0.2em}
\begin{itemize}
    \item We introduce a unified framework to quantify the total energy cost of learning, encompassing both model training and label acquisition, which is often overlooked in energy-efficient learning studies.
    \vspace{-0.3em}
    \item We conduct a comparative evaluation of supervised and self-supervised CL methods under varying data availability regimes, using energy consumption as a key evaluation metric.
    \item We provide practical insights into the trade-offs between labeling effort and energy efficiency, and offer guidelines for determining when supervised learning becomes energetically efficient or unfavorable.
    \vspace{-0.3em}
\end{itemize}

\section{Background and Motivation}
\label{sec:related_work}

Supervised learning has long been the dominant paradigm in AI/ML solutions, offering high predictive performance and relatively straightforward optimization pipelines. Nonetheless, in real-world deployments, models trained on static, fully-labeled datasets often fail to generalize due to misalignment between the training distribution and the observed data, i.e., leading to dataset shift problems.
These shifts can arise from seasonal changes in remote sensing imagery, sensor drift in wearable, or evolving environments in autonomous systems, 
to name a few examples. In such settings, retraining models with new, labeled data is both time-consuming and energy-intensive, posing a critical bottleneck for efficient and sustainable learning at the edge.

To address these challenges, particularly in dynamic data scenarios or when label data is scarce, CL methods have emerged as an effective alternative across different domains. Unlike supervised learning, which optimizes a \textit{cross-entropy} loss over labeled examples, CL methods typically rely on the \textit{InfoNCE} objective function \cite{infoNCE_oord2018representation}, which operates by pulling together representations of similar inputs (positive pairs) and pushing apart dissimilar ones (negative pairs), using pretext tasks like instance discrimination \cite{simclr_chen2020simple, supcon_khosla2020supervised}. More formally, given a batch of $N$ samples and their augmentations (indexed by $\mathcal{I} \equiv \left\{ 1,\dots,2N  \right\}$), \textit{InfoNCE} is typically estimated as
\begin{equation}
    \mathcal{L}_{self}= -\sum_{i \in \mathcal{I}}\text{log}\frac{\text{exp} \left( z_i \cdot z_{j(i)}/\tau \right)}{
    \sum_{a \in \mathcal{A}(i)} \text{exp} \left( z_i \cdot z_{a}/\tau \right)
    }
    \label{eq:infonce}
\end{equation}
where $j(i)$ indexes the other augmentation associated with the $i$th sample (also known as anchor), and $\mathcal{A}(i) \equiv \{\mathcal{I} \backslash i\}$ represents the set of all negative augmented samples.

The majority of CL approaches differ from each other in how they construct the positive $\mathcal{I}$ and negative sets $\mathcal{A}(i)$. For instance, SimCLR \cite{simclr_chen2020simple} uses a large batch size $N$ to ensure a diverse negative set. Unfortunately, having a large $\mathcal{A}(i)$ makes training memory-intensive, which limits its applicability in edge scenarios. 
 In contrast, MoCo \cite{he2020momentum} avoids the large-batch requirement by using a momentum encoder and a dynamic memory queue that maintains negative samples across training iterations.

Alternative approaches eliminate the need for explicit negatives entirely. 
Specifically, BYOL \cite{byol_grill2020bootstrap} and SimSiam \cite{simsiam_chen2021exploring} rely on dual-branch architectures and use, respectively, a momentum encoder and stop-gradient operation to prevent the model from collapsing into a trivial solution, where all embeddings converge to a constant vector. 
Although these alternatives may reduce memory usage, they often exhibit slower convergence rates, 
which may become problematic in edge scenarios where energy is limited and prolonged training directly translates to greater energy consumption.

To further improve sample efficiency, especially in low-label regimes, supervised and semi-supervised extensions of CL have also been proposed \cite{supcon_khosla2020supervised, zheng2021weakly,graf2021dissecting,barbano2022unbiased}. In supervised CL (SupCon) \cite{supcon_khosla2020supervised}, all samples sharing the same class label are treated as positives, enabling the use of multiple positive pairs per anchor  $\mathcal{P}(i) \subset \mathcal{A}(i)$.
The \textit{InfoNCE} objective (\ref{eq:infonce})  is then adapted to consider multiple positive samples per anchor by 
\begin{equation}
    \mathcal{L}_{self}^{(sup)}= 
    \sum_{i \in \mathcal{I}}\frac{-1}{\left| P(i) \right|}
    \sum_{p \in \mathcal{P}(i)}\text{log}\frac{\text{exp} \left( z_i \cdot z_{p}/\tau \right)}{
    \sum_{a \in \mathcal{A}(i)} \text{exp} \left( z_i \cdot z_{a}/\tau \right)
    }
    \label{eq:supcon_loss}
\end{equation}
where the augmented view $z_{j(i)}$, associated with the $i$th input, is now replaced by a set of embeddings  $z_p \in \mathcal{P}(i)$  corresponding to all samples that share the same class label as the $i$th anchor.

This strategy enhances intra-class cohesion in the latent space and generally yields superior performance compared to self-supervised approaches. Nonetheless, its reliance on labeled data reintroduces annotation costs, particularly problematic in edge settings where labeling is energy and/or resource-intensive.
To bridge the gap between fully supervised (\ref{eq:supcon_loss}) and unsupervised (\ref{eq:infonce}) extremes, semi-supervised contrastive learning frameworks, such as FixMatch, MixMatch, and more recent class-aware methods like CCSSL~\cite{seimcl_yang2022class}, leverage a labeled subset to guide the representation learning process while relying on pseudo-labeling or confidence-based heuristics for the unlabeled data.

Despite their promising results, the total energy footprint of these semi-supervised approaches, including the cost of training and labeling, remains underexplored. 
This work aims to fill that gap by providing a comprehensive, energy-aware evaluation of CL methods for edge deployments, where performance must be balanced with resource constraints. 
Through empirical analysis, we show that while SupCon may deliver superior accuracy, labeling incurs significantly higher energy consumption: over twice the energy cost of training the model. 
In contrast, self-supervised methods like SimCLR are more energy-efficient but less accurate. Semi-supervised techniques such as CCSSL offer a compelling trade-off, achieving near-supervised accuracy with significantly lower total energy\content{, e.g., requiring almost $4\times$ less energy to reach similar accuracies.}
These trends become even more pronounced in low-data regimes, where semi-supervised methods leverage partial labeling to maintain efficiency without compromising performance.

\section{Energy Consumption}
\label{sec:energy_consumption}

In what follows, we provide a detailed breakdown of the energy consumption associated with learning under different supervision regimes. Our analysis considers both: (i) the cost of labeling (both manual and with the help of an external server), often neglected in traditional evaluations, and (ii) the computational energy associated with training, including different components such as GPU, CPU, and RAM power. 
By accounting for both components, we aim to provide a comprehensive assessment of the energy footprint of learning pipelines under different supervision setups.
 
\subsection{Estimating Energy Cost of  Labeling}

{Manual annotation} remains a bottleneck in deploying supervised learning systems, particularly in scenarios where labels require expert knowledge (e.g., in medical domains) or large volumes of data must be labeled before training (e.g., in the context of signal processing and natural images). The energy cost associated with this process is often overlooked, yet it can constitute a significant portion of the total energy footprint. 

To estimate the energy consumed during manual labeling, we assume that labeling is performed on a standard desktop computer consuming $P_\mathrm{energy}$ watts (W) of power. Let $T_\mathrm{label} \in \mathbb{R}$ denote the average time (in seconds) to annotate a single data sample, and $K$ the total number of samples in the dataset. The total labeling energy is then approximated by:
\begin{equation}
\label{eq:energy_label}
E_\mathrm{labeling} = P_\mathrm{energy} \text{(Watts)} \times \frac{ K T_\mathrm{label} }{3600}  \text{(kWh)}.
\end{equation}

As a concrete example, let us consider the CIFAR-10 dataset, and the average time required to label a single natural image is approximately $T_\mathrm{label} = 10$ seconds \cite{chang2022duallabel}. Then, labeling the entire CIFAR-10 training set ($K = 50,000$ samples) would take around $139$ hours. Assuming that labeling is performed on a standard desktop computer consuming $P_\mathrm{energy} = 30$ watts of power, the total energy consumption required to label the entire training set may be estimated to be 
$30 \mathrm{Watts} \times 139 \mathrm{h}  = 4.17$~kWh.

While these energy values may appear modest, they represent a non-negligible fixed cost, especially in active learning or dynamic scenarios. Moreover, it is important to note that annotation time and energy requirements may vary considerably depending on dataset complexity, hardware considered, and 
transmission costs, 
which could further increase the energy expenditure associated with the labeling process.

\subsection{Estimating Energy Cost of Training}%Learning}
\label{sec:energy:learning}

We employ \texttt{CodeCarbon}  Python package \cite{codecarbon} %, bannour2021evaluating}
to measure energy consumption related to hardware operations, which allows us to measure energy consumption associated with the CPU, GPU, and memory. 
CodeCarbon measures the energy consumption of the CPU and GPU by %energy consumption associated with the first two (CPU and GPU) by 
periodically tracking (every $\Delta t = 15$ seconds) the power supply $P_\text{unit}(t_i), t_i = 1, \ldots, N$ of each processing unit. CPU energy consumption is tracked using the Intel Running Average Power Limit (RAPL) system management interface, whereas GPU energy consumption is monitored using the \textit{pynvml} Python library~\cite{bouza2023estimate}.
The total energy consumed by the processing units,
$$
E^{(k)}_\text{units} = \sum_{i=1}^N \left(P_\text{CPU}(t_i) + P_\text{GPU}(t_i)\right) \Delta t,
$$
depends on the duration of the training process ($N  \Delta t$) and the usage of the resources at each time $t_i = 1, \ldots, N$. 

CodeCarbon also allows us to track memory usage, for which we use the default \texttt{machine} mode, assuming an average memory power consumption of 
$37.5\times 10^{-2}$ W/Gb~\cite{maevsky2017evaluating_memory}. The process is similar to the above and periodically measures the resource consumption before converting it into energy consumption. Formally, 
$$
E^{(k)}_\text{memory} = %0.375
(37.5\times 10^{-2})\sum_{i=1}^N \Omega_\text{memory}(t_i) \Delta t,
$$
where $\Omega_\text{memory}(t_i)$ denotes the amount of memory (in Gb) consumed at each time $t_i = 1, \ldots, N$.

This approach allows us to estimate the accumulated energy consumed during the pre-processing and training stages from the energy consumed by the processing units (CPU and GPU) and the memory, as
$
E_\text{train}
=
\sum_{k=1}^K E^{(k)}_\text{units} + E^{(k)}_\text{memory}.
$
This methodology enables a consistent comparison of hardware energy demands across different 
learning frameworks.

\section{Numerical Results}
\label{sec:results}

\begin{figure}[t!]
    \centering
    \begin{subfigure}[t]{0.4\textwidth}
        \centering
\includegraphics[width=1\textwidth]{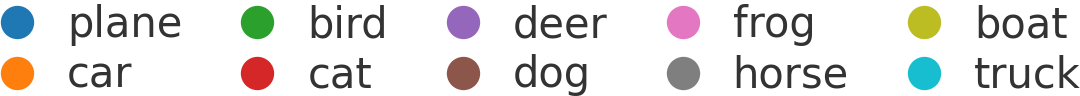}
    \end{subfigure}%
    \vspace{1em}
    
    \begin{subfigure}[t]{0.22\textwidth}  
   	 \centering  
        \includegraphics[width=0.8\textwidth]{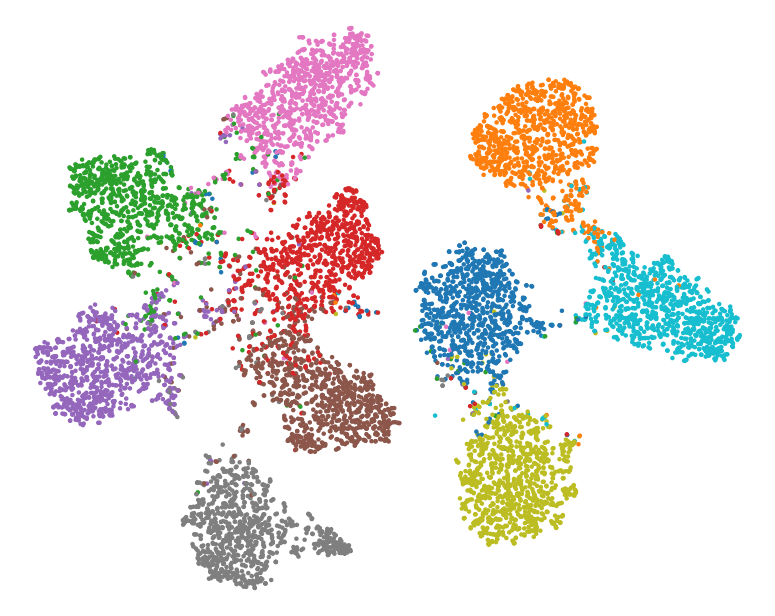}
        \caption*{\hspace{3pt}{Baseline}
        % \\ \hspace{5pt}(100\%)}
        }
    \end{subfigure}%
    \begin{subfigure}[t]{0.22\textwidth}
        \centering
        \includegraphics[width=0.8\textwidth]{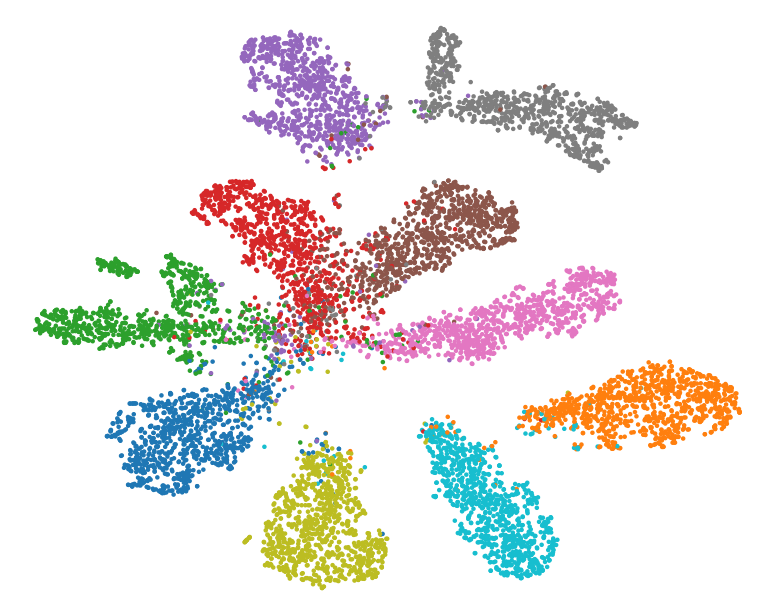}
        \caption*{{SupCon 
        % \\ (100\%)
        }}
    \end{subfigure}
    
    \begin{subfigure}[t]{0.22\textwidth}
        \centering
        \includegraphics[width=0.8\textwidth]{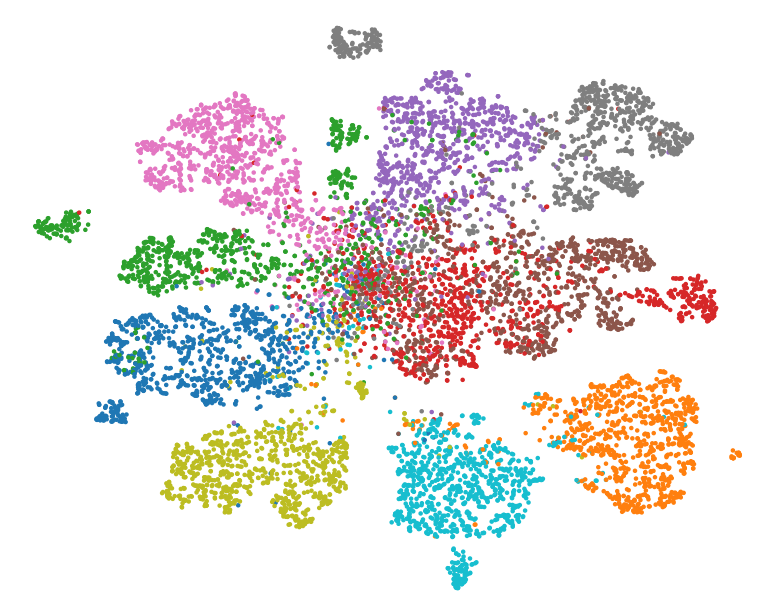}
        \caption*{{SimCLR 
        % \\ (100\%)
        }}
    \end{subfigure}% 
    \begin{subfigure}[t]{0.22\textwidth}
        \centering
        \includegraphics[width=0.8\textwidth]{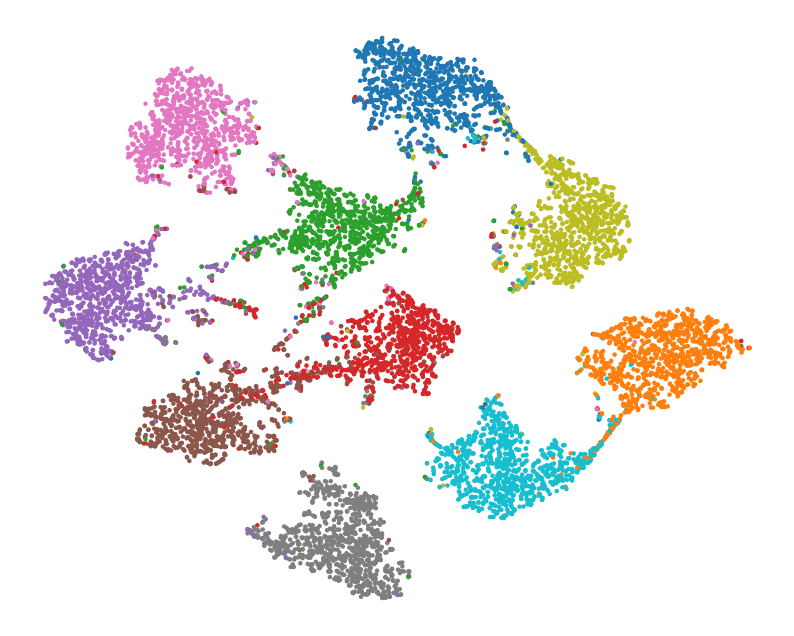}
        \caption*{CCSSL}
    \end{subfigure}%
    
    \caption{t-SNE visualization of learned feature spaces from each model. Colors represent different CIFAR-10 classes.  
    % \rmpp{TODO: labels on top}
    }
    \vspace{-1em}
    \label{fig:tsne} 
\end{figure}

In this section, we evaluate the total energy consumption and accuracy trade-offs of
vanilla cross-entropy and CL methods.
We consider four approaches: a standard supervised model trained with cross-entropy loss (baseline); SimCLR as a representative self-supervised CL method~\cite{simclr_chen2020simple}; SupCon as a supervised CL approach~\cite{supcon_khosla2020supervised}; and a semi-supervised CL model (CCSSL with FixMatch~\cite{seimcl_yang2022class}) that leverages partial label availability. Specifically, unless otherwise mentioned, we consider  50\% of the samples to be labeled and the remaining unlabeled.  This semi-supervised setting serves as a middle ground between fully labeled (cross-entropy) and fully unlabeled scenarios (SimCLR).  A visual comparison of the feature spaces learned by these models is presented in Fig.~\ref{fig:tsne}. We assess both performance and energy efficiency under varying levels of data availability.

\subsection{Experimental Setup}
\label{sec:results:setup}

 We employ ResNet-18 as the backbone architecture, yielding a 512-dimensional latent representation. All learning approaches--baseline, SimCLR, SupCon, and CCSSL\footnote{In our experiments, we analyzed a hybrid between SimCLR and SupCon by simply employing either (\ref{eq:infonce}) or (\ref{eq:supcon_loss}) depending on the label availability of the sample, but the model did not converge.
 % , so we consider CCSSL as semi-supervised learning representative despite consuming some energy for pseudo-labeling.
 }--are trained from scratch using this architecture.
With some abuse of notation, for CCSSL, we display the percentage of labeled samples of the training set by CCSSL$_\mathrm{(labeled)}$.
For SimCLR and CCSSL, we use a temperature parameter of $\tau = 0.1$, while for SupCon $\tau = 0.5$, following the setting in \cite{supcon_khosla2020supervised}. Each model is trained 
for $1,000$ epochs with a learning rate decay applied at epochs $700, 800$, and $900$. 
To assess model performance, we employ the kNN evaluation protocol \cite{knn_validation_wu2018unsupervised}, which provides a non-parametric estimate of accuracy based on similarity in the learned representation space.
Reported energy values are averaged across five independent runs with different random seeds.

Energy experiments are primarily conducted on the CIFAR-10 dataset, which consists of $60,000$ natural images of dimension $32 \times 32$ divided across $10$ classes, out of which $50,000$ images are used for training and $10,000$ for testing. 
We also consider the EuroSAT %\cite{eurosat_helber2019} 
dataset, which consists of $27,000$ Sentinel-2 satellite images of dimension $64 \times 64$. From these $27,000$ we randomly select $5,000$ to form the test set, and divide the remainder into training and validation sets. We note that the different dimensionality and number of samples of the two datasets will lead to different energy profiles and associated costs.

To simulate varying data availability scenarios, we evaluate all models under three training set sizes: $20\%$, $50\%$, and $100\%$ of the full dataset size. Each subset is sampled to maintain the original class distribution. The $20\%$ and $50\%$ scenarios emulate data-constrained environments, such as those often encountered in distributed or edge computing contexts, while the $100\%$ case represents a data-rich, centralized training scenario typical of high-capacity servers.

Finally,  as described in Section~\ref{sec:energy_consumption}, energy usage is monitored using \texttt{CodeCarbon}, tracking CPU, GPU, and memory consumption. All experiments run on a mid-range machine with an Intel i7 CPU (2.60GHz), 6GB RAM, and NVIDIA GeForce GTX 1660 Ti GPU—representative of research-scale or high-performance edge setups.

\subsection{Impact of Data Availability}%{Performance and Learning Cost Trade-off}
\label{sec:results:performance}

\begin{table*}[ht]
\centering
\caption{Test accuracy and estimated energy (averaged over five seeds) required to train  each model for $1,000$ epochs.}
\begin{tabular}{lr|cc|cc}
\toprule
\textbf{Model} & \textbf{\% Data} & \multicolumn{2}{c|}{\textbf{CIFAR-10}} & \multicolumn{2}{c}{\textbf{EuroSAT}} \\
& & \textbf{Test Accuracy (\%)} & \textbf{Energy (kWh)} & \textbf{Test Accuracy (\%)} & \textbf{Energy (kWh)} \\ 
\hline
\multirow{3}{*}{Baseline} 
& 20\%  & 84.00 & 0.26 & 73.24 & 0.41 \\ 
& 50\%  & 90.08 & 0.63 & 93.40 & 1.03 \\
& 100\% & 93.40 & 1.26 & 94.59 & 2.05 \\
\hline
\multirow{3}{*}{SimCLR} 
& 20\%  & 78.24 & 0.56 & 91.26 & 0.25 \\ 
& 50\%  & 82.36 & 1.26 & 94.84 & 0.61 \\
& 100\% & 90.36 & 2.67 & 97.06 & 1.14 \\
\hline
\multirow{3}{*}{SupCon} 
& 20\%  & 85.54 & 0.53 & 93.69 & 0.24 \\ 
& 50\%  & 92.15 & 1.25 & 96.71 & 0.61 \\
& 100\% & 94.37 & 2.51 & 97.92 & 1.14 \\
\hline
% \multirow{3}{*}{CCSSL (20/80)} 
% & 20\%  & -- & -- & -- & -- \\ 
% & 50\%  & 90.24 & -- & -- & -- \\
% & 100\% & -- & -- & -- & -- \\
% \hline
\multirow{3}{*}{CCSSL$_{(50)}$} 
& 20\%  & 85.48 & 0.45 & 94.37 & 0.46 \\ 
& 50\%  & 91.41 & 1.08 & 96.50 & 1.10 \\
& 100\% & 94.36 & 2.14 & 97.85 & 2.18 \\
\bottomrule
\end{tabular}
\label{tab:accuracy_energy}
\end{table*}

We start by evaluating the impact of data availability on both the performance and energy consumption of the four frameworks considered.
Table \ref{tab:accuracy_energy} presents the test accuracy and the estimated energy consumption (in kWh) required to train each model across the three data availability regimes.

The energy costs related to training are averaged over five different seeds, however, we refrain from displaying the associated variances as they were a few orders of magnitude smaller (approximately $10^{-3}$) than average values.
In terms of average energy consumed during training, the two CL methods (i.e., SimCLR and SupCon) consume comparable amounts of energy. This is expected as the main distinction between them lies in their respective definitions of positive and negative pairs.
This slight difference does not significantly affect their energy footprints, and similar patterns are observed across both the CIFAR-10 and EuroSAT datasets. 

A more prominent change is obtained when comparing contrastive methods against the supervised (baseline) and CCSSL approaches. Notably,  SupCon consistently outperforms all the other methods in terms of accuracy across all data regimes and datasets, highlighting the advantage of contrastive representation learning when labels are available. 
At full data availability and considering CIFAR-10, SupCon achieves a test accuracy of $94.37\%$, slightly surpassing the baseline ($93.40\%$) and the semi-supervised alternative ($94.36\%$) while significantly exceeding SimCLR ($90.36\%$). This, however, comes at the expense of consuming twice the energy of the baseline.

We also note that the benefit of having access to labeling increases as the amount of available training data decreases (50\% and 20\% scenarios). For example, on the CIFAR-10 dataset with only 50\% of the data, SupCon improves in $2.07\%$ the accuracy over the baseline, while consuming an additional $0.62$ kWh. In the same scenario, using CCSSL yields a $1.33\%$ accuracy gain while incurring an additional energy cost of $0.45$ kWh during training. To contextualize these values, the energy consumed corresponds to charging a typical smartphone $\sim$ 30 times for SupCon and $\sim$ 20 times for CCSSL.

The baseline model trained with cross-entropy loss reaches a higher test accuracy of $93.40\%$ using the entire CIFAR-10 dataset, while consuming approximately half the energy required by SimCLR, which reaches $90.36\%$ under the same conditions.  
Furthermore, it is worth noting that under limited data availability, the baseline approach remains competitive. When trained with only $50\%$ of the data, it reaches comparable accuracy ($90.08\%$) to SimCLR ($90.36\%$) trained on the entire CIFAR-10 dataset, while consuming just one-quarter of the energy.
Moreover, when considering EuroSAT, going from using 100\%  of the training data to 50\% leads to an equivalent saving in the energy required for training, but with a small reduction of only 1.09\% in accuracy.

\subsection{ Impact of Labeling}% in Energy Profiling}
\label{sec:results:label}

\begin{figure}[t!]
\centering
\includegraphics[width=0.9\linewidth]{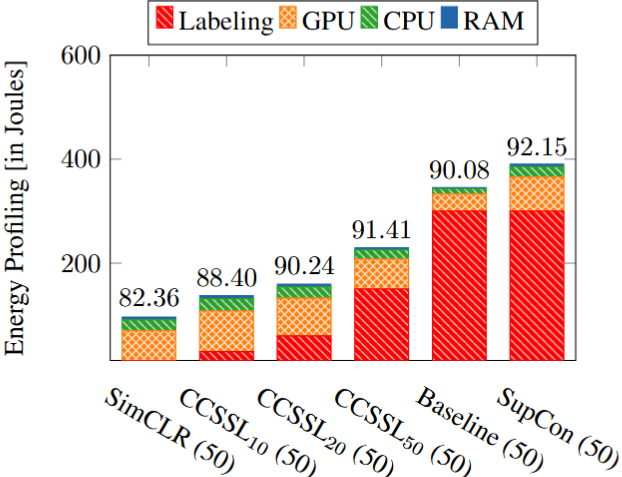}
 
     \caption{
      Energy profiling 
      during training using the CIFAR-10 dataset 
      decomposed by component: GPU, CPU, RAM, and labeling. SupCon and Baseline include the estimated labeling (in red) cost of  $\sim300$ Joules per sample. CCSSL's subscript denotes the percentage of labeled samples considered.
     }
    % \vspace{-1pt}
        \label{fig:energy_profiling}
    \vspace{-1\baselineskip}
\end{figure}

Our analysis has thus far solely focused on
the accumulated energy during training. 
However, to fully assess the energy costs of supervised learning algorithms, we must also account for the energy expended during data labeling. We thus extend our analysis to incorporate the energy spent on labeling.
Fig. \ref{fig:energy_profiling} displays similar results as above,  now breaking down energy usage by hardware (GPU, CPU, and RAM) and including the cost of labeling a single sample.
\content{We consider the low-data regime and all models are trained on 50\% of the CIFAR-10 training set.
The subscript in CCSSL$_\mathrm{(X)}$ indicates the portion of labeled data. For instance, CCSSL$_{(20)}$ (50) refers to CCSSL using $50\%$ of the training set, with 20\% labeled and $80\%$  unlabeled. 
}

\content{
Particularly, we note that SimCLR becomes the most energy-efficient approach, while the baseline and SupCon are the most energy consuming ones. Specifically, while these models achieve high predictive accuracy, relying on labeled data introduces a substantial fixed energy overhead. As illustrated in Fig. \ref{fig:energy_profiling}, this labeling cost (in red) often exceeds the total hardware-related (blue, green, and orange) energy consumption for these supervised methods. As suggested from the discussion above, this labeling cost can be controlled when employing semi-supervised learning solutions. For instance, CCSSL$_{50}$ (50) reaches very similar accuracies as SupCon, i.e., 91.41\% and 92.15\% respectively, while consuming $2.4\times$ the energy of SimCLR (for reference, SupCon consumes $4\times$ more SimCLR's energy).  
Similarly, CCSSL$_{10}$  (50)  consumes only $1.4\times$ more energy than SimCLR while delivering an improvement in accuracy of $6.04\%$.
These results further illustrate that the recurring expenditure of a large amount of energy per sample can prove energetically prohibitive and that there exists a trade-off between performance, labeling costs, and total energy consumption. 
}

\section{Conclusion}
\label{sec:conclusion}

In this work, we presented a comprehensive analysis of the energy-performance trade-offs in contrastive learning methods with a focus on edge scenarios. By measuring both training energy and the often-overlooked cost of labeling, we uncover critical trade-offs between performance and resource consumption. Our empirical analysis reveals that while supervised contrastive methods such as SupCon achieve the highest accuracy, their overall energy footprint is dominated by the cost of labeling. 
On the other hand, self-supervised approaches like SimCLR are more energy-efficient but fall short in performance. Semi-supervised approaches, particularly CCSSL, strike a practical balance, achieving high accuracy with significantly lower energy demands, especially in low-label settings. These insights underscore the importance of energy-aware benchmarking and highlight the viability of semi-supervised learning for sustainable, scalable AI at the edge. Future work could further investigate how these trade-offs behave in distributed learning settings, specifically when the data labeling is unbalanced.

\bibliographystyle{IEEEbib}
\bibliography{refs}

\end{document}